\pdfoutput=1
\PassOptionsToPackage{table}{xcolor}
\documentclass[11pt]{article}
\usepackage[preprint]{acl}
\usepackage{times}
\usepackage[T1]{fontenc}
\usepackage[utf8]{inputenc}
\usepackage{latexsym}
\usepackage{amsmath,amssymb,amsthm}
\usepackage{booktabs}
\usepackage{graphicx}
\usepackage{microtype}
\usepackage{subcaption}
\usepackage{multirow}
\usepackage[most]{tcolorbox}
\usepackage[ruled,vlined,linesnumbered]{algorithm2e}
\definecolor{ourgreen}{RGB}{230,245,230}
\usepackage{enumitem}
\usepackage{adjustbox}
\usepackage{fancyvrb}
\usepackage{tabularx}
\usepackage{array}
\usepackage{makecell}
\usepackage{pifont}
\usepackage{graphicx}
\usepackage{tabularx}
\usepackage{booktabs}
\usepackage{makecell}
\newcommand{\rwcheck}{\textcolor{green!55!black}{\ding{51}}}
\newcommand{\rwxmark}{\textcolor{red!70!black}{\ding{55}}}
\newtheorem{proposition}{Proposition}
\newtheorem{assumption}{Assumption}

\tcbuselibrary{skins,breakable}
\newtcolorbox{templatebox}[2][]{
	enhanced, breakable,
	colback=#2!5!white,
	colframe=#2!70!black,
	fonttitle=\bfseries\small,
	title={#1},
	left=4pt, right=4pt, top=4pt, bottom=4pt,
	boxrule=0.8pt,
}
\colorlet{SystemColor}{blue}
\colorlet{UserColor}{teal}
\colorlet{SkeletonColor}{orange}
\colorlet{CorrectionColor}{red}
\colorlet{ResponseColor}{purple}

\graphicspath{{./figures/}}

\title{Agentic vs. Parametric World Models:\\
	Hybrid Planning for Reliable Language Agents}

\author{\textbf{Xinyuan Song}$^{1}$ \quad
	\textbf{Zekun Cai}$^{2}$ \\
	$^{1}$Emory University, Atlanta, GA, USA \quad
	$^{2}$The University of Tokyo, Tokyo, Japan \\
	\texttt{xinyuan.song@emory.edu, caizekun@csis.u-tokyo.ac.jp} \\
}
\date{}

\newcommand{\method}{\textsc{Hybrid-WM}}
\begin{document}
	\maketitle
	
	\begin{abstract}
		Language agents, such as ReAct, Reflexion, and RAP, implicitly predict how the world will change as they plan and act, using imagined state transitions to guide their decisions. These predictions are encoded in the agents' backbone LLMs as internal beliefs about the environment, effectively forming an implicit world model for multi-step planning. However, hallucinated state information can become embedded in this world model, leading to persistent inaccuracies that influence subsequent decisions. Traditional parametric world models, such as Dreamer and MuZero, learn environment dynamics through explicit transition models, but their limited model capacity makes them difficult to apply directly to modern language-agent planning. We study this tradeoff in graph-structured planning environments and seek to combine the strengths of both paradigms. We propose Hybrid World-Model Planning (\method{}), which preserves the language agent's semantic planning capabilities while leveraging a lightweight parametric world model for more accurate transition predictions. Specifically, the parametric model constructs a compact planning skeleton to guide action generation while maintaining consistency between the agent's imagined transitions and the model's predictions. Experiments across four graph-structured planning benchmarks demonstrate that \method{} consistently improves task success and reduces hallucinated-state propagation with modest computational overhead. These findings highlight the effectiveness of combining semantic planning with lightweight parametric grounding for reliable long-horizon language-agent planning. Our code is available at:
		\url{https://github.com/Hik289/Agent-vs-param.git}.
	\end{abstract}

	\section{Introduction}
	
	\label{sec:intro}
	
	\begin{figure*}[!ht]
		\centering
		\includegraphics[width=0.85\textwidth]{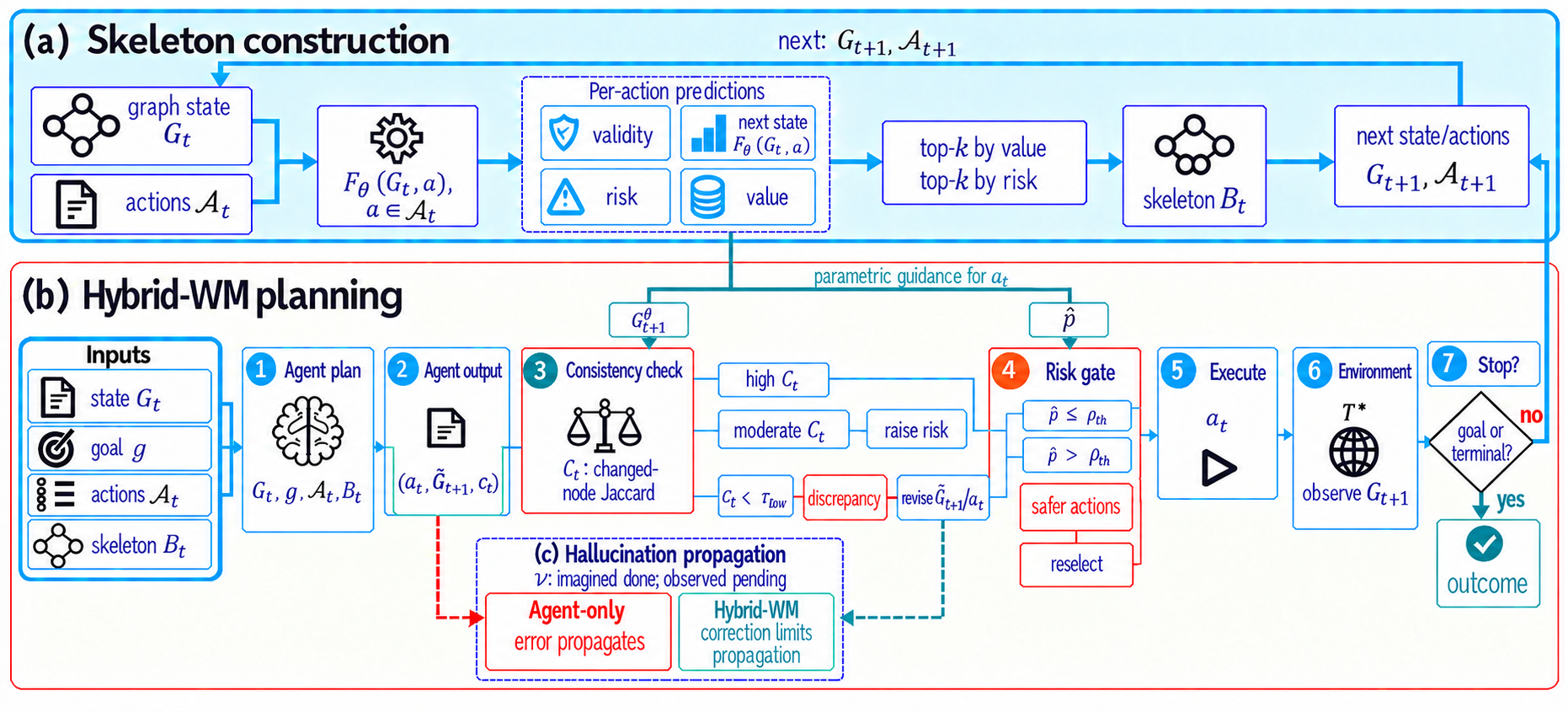}
		
		\caption{\textbf{Overview of Hybrid World-Model Planning (\method{}).}
			\textbf{(a) Skeleton construction:} A parametric world model predicts action outcomes and constructs a compact planning skeleton.
			\textbf{(b) Closed-loop planning:} The LLM plans using the skeleton, with transition consistency and risk-aware control guiding execution.
			\textbf{(c) Hallucination propagation:} \method{} limits the persistence of incorrect imagined states across decisions.}
		\label{fig:hybridwm-framework}
		
	\end{figure*}
	
	Language agents, such as ReAct~\citep{yao2022react}, Reflexion~\citep{shinn2023reflexion}, and RAP~\citep{hao2023reasoning}, plan by predicting how their actions will change the environment and using these imagined state transitions to guide subsequent decisions~\citep{yao2023tree}. Specifically, agents reason about their current state, select actions, and predict the resulting changes, which inform subsequent planning steps. This process supports multi-step interaction across embodied environments such as ALFWorld~\citep{shridhar2021alfworld}, interactive shopping tasks such as WebShop~\citep{yao2022webshop}, and web navigation environments such as WebArena~\citep{zhou2024webarena}. These predicted state changes form internal beliefs about the environment, effectively serving as an \emph{implicit world model} for multi-step planning. Rather than relying on an explicit transition model, language agents encode and update these beliefs within their backbone large language models (LLMs), leveraging semantic reasoning to predict future states and select actions.

	However, relying on an implicit world model introduces a fundamental reliability challenge. Because language agents use their backbone LLMs to predict state changes, these predictions may contain hallucinated information that does not match the actual environment~\citep{ji2023survey,huang2023look}. For example, an agent may predict that a door has opened after issuing an action, even though the door remains closed. If the agent does not verify this prediction, it may continue planning as if the door were open and attempt to enter the next room. When this action fails, the agent may still rely on its earlier incorrect prediction, leading to further planning errors. In this way, a single hallucinated state can affect multiple decisions and become increasingly difficult to correct as the planning horizon grows~\citep{shinn2023reflexion,yao2023tree}. Although hallucination has been widely studied in language generation~\citep{ji2023survey} and reasoning faithfulness~\citep{lyu2023faithfulcot}, how incorrect state predictions persist and affect later decisions in language-agent planning remains insufficiently understood.
	
	Traditional parametric world models offer a different approach to representing environment dynamics. Beginning with Dyna~\citep{sutton1991dyna}, model-based reinforcement learning has used learned transition functions to predict future states and support planning. More recent approaches, including Dreamer~\citep{hafner2020dreamer}, DreamerV3~\citep{hafner2023dreamerv3}, and MuZero~\citep{schrittwieser2020muzero}, demonstrate the effectiveness of learning compact representations of environment dynamics for sequential decision-making. Unlike implicit language-based world models, these parametric models provide explicit predictions that can be trained and evaluated against observed transitions. However, parametric world models, such as Dreamer~\citep{hafner2023dreamerv3} and MuZero~\citep{schrittwieser2020muzero}, are primarily designed to learn environment dynamics within structured state and action spaces. This makes them difficult to apply directly to language-agent planning, where interpreting natural-language goals and determining appropriate actions require the semantic reasoning capabilities of LLMs~\citep{hao2023reasoning}. Consequently, parametric world models are better suited to supporting transition prediction than replacing LLM-based planning.
	
	These observations reveal a fundamental tradeoff between two forms of world modeling. Language agents provide flexible semantic reasoning but may accumulate hallucinated beliefs~\citep{huang2023look,hao2023reasoning}, whereas parametric world models offer explicit transition predictions but lack comparable semantic flexibility~\citep{hafner2023dreamerv3,schrittwieser2020muzero}. Rather than replacing one with the other, we investigate whether their complementary strengths can be combined to improve planning reliability. We study this question in graph-structured planning environments~\citep{liu2023llm+p,guan2023leveraging,silver2024generalized}, where state transitions and action dependencies can be explicitly evaluated while preserving the semantic reasoning demands of multi-step tasks. This setting allows us to examine not only whether an agent completes its goal, but also whether its imagined environment states remain faithful throughout the planning process.
	
	To address this challenge, we propose \textbf{Hybrid World-Model Planning} (\method{}), a framework that combines an LLM planner with a lightweight parametric world model. As illustrated in Figure~\ref{fig:hybridwm-framework}, \method{} preserves the language agent as the primary reasoning engine while using parametric transition predictions to ground its decisions. Specifically, the parametric model constructs a compact planning skeleton that guides action generation, while the framework maintains consistency between the agent's imagined transitions and the model's predictions. A risk-aware mechanism further constrains action selection before execution, and environment feedback updates the planning context for subsequent decisions. By combining semantic reasoning with structured transition grounding, \method{} reduces hallucinated-state propagation without replacing the language agent's planning capabilities.
	
	We evaluate \method{} across four graph-structured planning benchmarks: TaskGraph for workflow planning~\citep{liu2023llm+p}, ToolChain for tool-use dependencies~\citep{qin2023toolllm}, ResourceAlloc for resource allocation, and RepairFlow for failure recovery~\citep{guan2023leveraging}. We compare language-agent planners, standalone parametric world models, and their hybrid combination. Beyond task success, we evaluate the accuracy of predicted state transitions and how prediction errors affect subsequent decisions. Our experiments show that \method{} consistently improves planning success and reduces error propagation with modest computational overhead. These results demonstrate that lightweight parametric world models can ground language-agent planning while preserving the semantic reasoning capabilities of LLMs.
	
	Our contributions are summarized as follows:
	\begin{itemize}[left=0em]
		\item We study the complementary strengths and limitations of implicit LLM-based world models and explicit parametric world models for multi-step planning.
		\item We investigate how incorrect state predictions persist and affect subsequent decisions, introducing metrics to measure transition accuracy and error propagation over planning horizons.
		\item We propose \method{}, a hybrid planning framework that combines a lightweight parametric planning skeleton with LLM-based semantic reasoning through transition consistency checks and risk-aware action selection.
		\item We evaluate \method{} across four graph-structured planning benchmarks, demonstrating consistent improvements in task success and transition accuracy, reduced error propagation, and modest computational overhead.
	\end{itemize}

	\begin{table*}[!ht]
		\centering
		\caption{\textbf{Comparison with closely related world-model agents.}
			\textbf{Learned WM}: learned environment dynamics;
			\textbf{Explicit state}: structured environment-state representation;
			\textbf{LLM planner}: LLM-based action generation;
			\textbf{Risk-aware}: explicit action-risk estimation and filtering;
			\textbf{Consistency}: comparison of independent transition predictions;
			\textbf{Correction}: disagreement-triggered same-step revision.
			\rwcheck\ denotes presence; \rwxmark\ denotes absence.}
		\label{tab:rw-novelty}
		
		\footnotesize
		\setlength{\tabcolsep}{5pt}
		\renewcommand{\arraystretch}{1.18}
		
		\resizebox{\textwidth}{!}{%
			\begin{tabular}{
					@{}
					>{\raggedright\arraybackslash}p{5.5cm}
					>{\raggedright\arraybackslash}p{2.2cm}
					*{4}{>{\centering\arraybackslash}p{1.35cm}}
					>{\centering\arraybackslash}p{1.65cm}
					@{\hspace{5pt}}
					>{\centering\arraybackslash}p{1.65cm}
					@{}
				}
				\toprule
				\textbf{Work}
				& \textbf{WM form}
				& \makecell{\textbf{Learned}\\\textbf{WM}}
				& \makecell{\textbf{Explicit}\\\textbf{state}}
				& \makecell{\textbf{LLM}\\\textbf{planner}}
				& \makecell{\textbf{Risk-}\\\textbf{aware}}
				& \textbf{Consistency}
				& \textbf{Correction} \\
				\midrule
				
				\multicolumn{8}{@{}l}{
					\textit{Executable, symbolic, and knowledge-based world models}} \\
				
				WorldCoder~\citep{tang2024worldcoder}
				& Executable code
				& \rwcheck & \rwcheck & \rwcheck & \rwxmark & \rwxmark & \rwxmark \\
				
				WKM~\citep{qiao2024wkm}
				& Parametric LLM
				& \rwcheck & \rwcheck & \rwcheck & \rwxmark & \rwxmark & \rwxmark \\
				
				WALL-E~\citep{zhou2025walle2}
				& LLM + rules
				& \rwcheck & \rwcheck & \rwcheck & \rwxmark & \rwxmark & \rwxmark \\
				
				\midrule
				\multicolumn{8}{@{}l}{
					\textit{Neural and LLM-based transition models}} \\
				
				Causal WM~\citep{gkountouras2025causalwm}
				& Causal neural
				& \rwcheck & \rwcheck & \rwxmark & \rwxmark & \rwxmark & \rwxmark \\
				
				WMA~\citep{chae2025webagentswm}
				& Trained LLM
				& \rwcheck & \rwcheck & \rwcheck & \rwxmark & \rwxmark & \rwxmark \\
				
				WebDreamer~\citep{gu2024llmworldmodel}
				& Prompted LLM
				& \rwxmark & \rwxmark & \rwcheck & \rwxmark & \rwxmark & \rwxmark \\
				
				Precondition--Effect WM~\citep{xie2025preconditioneffect}
				& Two tuned LLMs
				& \rwcheck & \rwcheck & \rwcheck & \rwxmark & \rwxmark & \rwxmark \\
				
				From Word to World~\citep{li2026wordtoworld}
				& Trained LLM
				& \rwcheck & \rwcheck & \rwxmark & \rwxmark & \rwxmark & \rwxmark \\
				
				R-WoM~\citep{mei2026rwom}
				& LLM + retrieval
				& \rwxmark & \rwcheck & \rwcheck & \rwxmark & \rwxmark & \rwxmark \\
				
				\midrule
				
				\rowcolor{blue!9}
				\textbf{Hybrid-WM (ours)}
				& \textbf{Parametric + LLM}
				& \rwcheck
				& \rwcheck
				& \rwcheck
				& \rwcheck
				& \rwcheck
				& \rwcheck \\
				
				\bottomrule
			\end{tabular}%
		}
		
	\end{table*}

	\section{Related Work}
	
	\label{sec:related}
	
	\paragraph{LLM-based planning and world models.}
	LLMs increasingly serve as implicit world models for multi-step planning. ReAct~\citep{yao2022react} interleaves reasoning and action, Reflexion~\citep{shinn2023reflexion} incorporates feedback, and Plan-and-Solve~\citep{wang2023plan} decomposes tasks. Tree of Thoughts~\citep{yao2023tree} and Graph of Thoughts~\citep{besta2024graph} introduce structured search, while LATS~\citep{liu2023lats} and RAP~\citep{hao2023reasoning} combine language-based predictions with planning. These methods support interactive tasks in ALFWorld~\citep{shridhar2021alfworld}, WebShop~\citep{yao2022webshop}, WebArena~\citep{zhou2024webarena}, and $\tau$-bench~\citep{yao2024taubench}. However, incorrect state predictions may persist across planning steps. Instruction tuning~\citep{ouyang2022instructgpt} and agent specialization~\citep{zeng2023agenttuning,chen2023fireact} improve agent capabilities but do not guarantee transition accuracy. In contrast, \method{} grounds LLM predictions in an independently learned world model.
	
	\paragraph{Parametric and structured world models.}
	Parametric world models learn environment dynamics for model-based planning, from Dyna~\citep{sutton1991dyna} and probabilistic ensembles~\citep{chua2018pets} to Dreamer~\citep{hafner2020dreamer,hafner2023dreamerv3} and MuZero~\citep{schrittwieser2020muzero}. Graph-based architectures, including GCN~\citep{kipf2017gcn}, GraphSAGE~\citep{hamilton2017graphsage}, GAT~\citep{velickovic2018gat}, and graph transformers~\citep{rampasek2022gps}, further support relational state representations. While these models provide structured transition predictions, they lack the semantic flexibility of LLM planners. Recent approaches explore executable world models in WorldCoder~\citep{tang2024worldcoder}, symbolic rules in WALL-E~\citep{zhou2024walle}, and LLM-based environment modeling~\citep{gu2024llmworldmodel}. Unlike these methods, \method{} uses lightweight parametric predictions of action validity, state changes, risk, and value to guide LLM planning through a compact planning skeleton.
	
	\paragraph{Hybrid planning and hallucination correction.}
	Existing methods address LLM hallucination~\citep{ji2023survey} through self-consistency~\citep{wang2022selfconsistency}, chain-of-verification~\citep{dhuliawala2023chainverify}, tool-assisted critique~\citep{gou2024critic}, and process reward models~\citep{lightman2023letsverify}. Hybrid approaches introduce structured planning through LLM+P~\citep{liu2023llm+p}, PDDL-based world models~\citep{guan2023leveraging}, and LLM-Modulo~\citep{kambhampati2024llms}, while structured decoding~\citep{geng2023grammar} constrains output validity. However, these methods do not directly address how incorrect state predictions affect subsequent planning decisions. \method{} combines skeleton-guided action generation with transition consistency checks and risk-aware selection, enabling targeted correction before execution and reducing error propagation across planning steps.
	
	\paragraph{Comparison with prior work.}
	Table~\ref{tab:rw-novelty} compares \method{} with existing world-model agents in terms of representation, planning, and correction mechanisms. Unlike prior approaches, \method{} integrates lightweight parametric predictions with LLM-based planning to support transition consistency checking, targeted correction, and risk-aware action selection.

	
	\section{Method}
	\label{sec:method}

	\subsection{Problem Formulation}
	\label{sec:problem}
	
	We formulate multi-step language-agent planning as sequential decision-making guided by predicted environment transitions. We distinguish the agent's imagined state changes from actual environment outcomes and assess prediction accuracy, error propagation, and computational cost.

	\paragraph{Multi-step Language Planning}
	
	We consider a multi-step planning task in which an agent interacts with an environment to achieve a natural-language goal $g$ within a horizon $H$. At each step $t$, the environment is represented as a typed graph:
	\begin{equation}
	\label{eq:graph_state}
	G_t=(V_t,E_t,X_t,Z_t),
	\end{equation}
	where $V_t$ denotes the set of entities, $E_t$ captures their dependencies, $X_t$ records node attributes and statuses, and $Z_t$ identifies goal nodes. Each node has one of five statuses: \textsc{pending}, \textsc{active}, \textsc{completed}, \textsc{failed}, or \textsc{skipped}.
	
	Given the current state $G_t$ in Eq.~\eqref{eq:graph_state}, the goal $g$, and available actions $\mathcal{A}_t$, the agent selects an action $a_t\in\mathcal{A}_t$ and imagines the resulting state $\widetilde{G}_{t+1}$. The environment then transitions according to
	\begin{equation}
	\label{eq:true_transition}
	G_{t+1}=T^\star(G_t,a_t),
	\end{equation}
	where $T^\star$ is the true environment transition function. The task succeeds when all goal nodes reach \textsc{completed} within $H$ steps. The difference between the imagined state $\widetilde{G}_{t+1}$ and the actual state $G_{t+1}$ in Eq.~\eqref{eq:true_transition} characterizes the agent's world-model error, which may influence subsequent planning decisions.

	\paragraph{Agent World-Model Error}
	
	The agent's imagined next state $\widetilde{G}_{t+1}$ may differ from the actual state $G_{t+1}$ in Eq.~\eqref{eq:true_transition}. We define a \emph{hallucinated state} as an incorrectly predicted node status and measure such errors using the \emph{Hallucinated-State Rate (HSR)}, the fraction of predicted node statuses that disagree with the actual state.
	
	Over a planning horizon $H$, let $\mathcal{E}_k$ denote the event of a hallucinated state or invalid action at step $k$. Under stepwise independence, the probability of at least one error is
	\begin{equation}
	\label{eq:exact}
	\Pr\left(\bigcup_{k=1}^{H}\mathcal{E}_k\right)=1-\prod_{k=1}^{H}\left(1-\Pr(\mathcal{E}_k)\right).
	\end{equation}
	Since incorrect predictions may influence subsequent decisions, stepwise errors need not be independent. Equation~\eqref{eq:exact} therefore provides an independence-based reference rather than a general bound.

	\paragraph{Parametric World-Model Error}
	
	We consider a lightweight parametric world model $F_\theta$ that predicts the next environment state from the current graph $G_t$ and action $a_t$:
	\begin{equation}
	\label{eq:parametric_world_model}
	G_{t+1}^{\theta}=F_\theta(G_t,a_t).
	\end{equation}
	We train $F_\theta$ using transitions collected from the true environment function $T^\star$ in Eq.~\eqref{eq:true_transition}. For each observed transition, the model predicts the next graph state and is optimized to minimize the discrepancy between its prediction $G_{t+1}^{\theta}$ and the actual state $G_{t+1}$, with outside supervision for action validity, reward, termination, and risk. We evaluate prediction accuracy using node-level mean squared error (NodeMSE). Unlike the agent's imagined state $\widetilde{G}_{t+1}$, the parametric model learns directly from observed environment transitions, providing complementary guidance for multi-step planning.

	\subsection{Hybrid World-Model Planning}

	We propose Hybrid World-Model Planning (\method{}), which combines agent-based semantic reasoning with lightweight parametric transition predictions. As illustrated in Figure~\ref{fig:hybridwm-framework}, \method{} integrates planning skeleton construction, agent planning, transition consistency checking, and risk-aware action selection into a closed-loop framework. The complete procedure is presented in Algorithm~\ref{alg:hybridwm}.

	\paragraph{Planning Skeleton Construction}

	Given the current graph state $G_t$ and candidate actions $\mathcal{A}_t$, the parametric world model $F_\theta$ predicts the next state $F_\theta(G_t,a)$ for each $a\in\mathcal{A}_t$, following Eq.~\eqref{eq:parametric_world_model}. It also estimates action validity, risk, and future planning value. As illustrated in Figure~\ref{fig:hybridwm-framework}(a), we construct a compact planning skeleton by retaining the top-$k$ actions ranked by predicted value and the top-$k$ ranked by predicted risk:
	\begin{equation}
	\label{eq:planning_skeleton}
	B_t=\{(a,F_\theta(G_t,a)):a\in\mathcal{A}_t\}.
	\end{equation}
	The planning skeleton $B_t$ in Eq.~\eqref{eq:planning_skeleton} associates each candidate action with its predicted next state, following Eq.~\eqref{eq:parametric_world_model}. These predictions provide structured guidance for action selection and serve as a reference for checking the agent's imagined state $\widetilde{G}_{t+1}$, complementing its semantic reasoning.

	\paragraph{Agent Planning.}
	As illustrated in Figure~\ref{fig:hybridwm-framework}(b), the agent receives the current state $G_t$, goal $g$, available actions $\mathcal{A}_t$, and planning skeleton $B_t$ defined in Eq.~\eqref{eq:planning_skeleton}. It reasons about the goal, current node statuses, and action dependencies to select an action $a_t$ and imagine the resulting state $\widetilde{G}_{t+1}$ with confidence $c_t$. The skeleton provides predicted transitions from Eq.~\eqref{eq:parametric_world_model}, helping the agent assess action consequences and identify feasible paths toward the goal. The agent uses these predictions as guidance while retaining the flexibility to make decisions based on semantic constraints and task requirements.

	\paragraph{Transition Consistency and Correction.}
	As illustrated in Figure~\ref{fig:hybridwm-framework}(b), we compare the agent's imagined next state $\widetilde{G}_{t+1}$ with the parametric prediction $G_{t+1}^{\theta}$ from Eq.~\eqref{eq:parametric_world_model}. We measure their agreement using Jaccard similarity~\citep{jaccard1912distribution} over nodes whose statuses are predicted to change relative to $G_t$:
	\begin{equation}
	\label{eq:transition_consistency}
	\small
	C_t=\frac{|\{v:\widetilde{G}_{t+1}[v]\neq G_t[v]\}\cap\{v:G_{t+1}^{\theta}[v]\neq G_t[v]\}|}{|\{v:\widetilde{G}_{t+1}[v]\neq G_t[v]\}\cup\{v:G_{t+1}^{\theta}[v]\neq G_t[v]\}|}.
	\end{equation}
	We set $C_t=1$ when neither model predicts a change. Based on Eq.~\eqref{eq:transition_consistency}, high consistency allows the agent to retain its selected action, moderate consistency increases its estimated risk, and low consistency triggers revision. During revision, the parametric prediction $G_{t+1}^{\theta}$ provides feedback on discrepancies with the imagined state $\widetilde{G}_{t+1}$, guiding the agent to correct its prediction or reconsider its action.
	
	\paragraph{Agent Execution.}
	When low consistency triggers correction, the parametric prediction $G_{t+1}^{\theta}$ guides the agent to revise its imagined state $\widetilde{G}_{t+1}$ or selected action $a_t$, as measured by Eq.~\eqref{eq:transition_consistency}. The agent then evaluates the predicted action risk and considers lower-risk alternatives from $\mathcal{A}_t$ when necessary. The final action is executed according to Eq.~\eqref{eq:true_transition}, and the observed state $G_{t+1}$ informs the next planning step, completing the closed-loop process.
	
	\begin{algorithm}[!ht]
		\SetAlgoLined
		\KwIn{Initial state $G_0$, goal $g$, parametric world model $F_\theta$, horizon $H$}
		\For{$t=0,\ldots,H-1$}{
			Obtain available actions $\mathcal{A}_t$\;
			Construct planning skeleton $B_t$ using Eq.~\eqref{eq:planning_skeleton}\;
			Select action $a_t$ and imagine next state $\widetilde{G}_{t+1}$ guided by $B_t$\;
			Predict $G_{t+1}^{\theta}$ using Eq.~\eqref{eq:parametric_world_model}\;
			Evaluate transition consistency $C_t$ using Eq.~\eqref{eq:transition_consistency}\;
			Revise the imagined state or action when consistency is low, and adjust action risk when consistency is moderate\;
			Reconsider high-risk actions using safer alternatives from $\mathcal{A}_t$\;
			Execute $a_t$ and observe $G_{t+1}$ using Eq.~\eqref{eq:true_transition}\;
			Stop if the goal is achieved or the episode terminates\;
		}
		\caption{Hybrid World-Model Planning (\method{}).}
		\label{alg:hybridwm}
	\end{algorithm}

	\section{Experiments}
	\label{sec:experiments}
	
	\paragraph{Experimental Setup.}
	We evaluate \method{} on four graph-structured planning benchmarks: TaskGraph (workflow dependencies)~\citep{liu2023llm+p}, ToolChain (tool-use dependencies)~\citep{qin2023toolllm}, ResourceAlloc (resource contention), and RepairFlow (cascading failures)~\citep{guan2023leveraging}. These environments capture challenges in embodied planning~\citep{shridhar2021alfworld}, interactive decision-making~\citep{yao2022webshop}, general agent evaluation~\citep{liu2023agentbench}, and web-based planning~\citep{zhou2024webarena}. Each benchmark includes training, validation, and test tasks, with both in-distribution and out-of-distribution evaluation. We train six parametric world-model backbones: MLP~\citep{rumelhart1986learning}, GCN~\citep{kipf2017gcn}, MPNN~\citep{gilmer2017mpnn}, GPS~\citep{rampasek2022gps}, and two task-specific variants, ActionNode and ErrorAware, using oracle transitions and compare agentic-only, parametric-only, and hybrid planners. To enable large-scale evaluation, we use a behavioral simulator calibrated against real GPT-4o-mini~\citep{openai2024gpt4omini} executions while preserving ground-truth graph transitions. The simulator reproduces agent action selection, hallucinated state predictions, correction outcomes, and inference costs. All metrics are computed from shared per-step trajectories. Detailed benchmark specifications, model architectures, training procedures, and simulator calibration are provided in Appendix~\ref{app:experimental_details}.

	\subsection{Main Comparison}
	
	Table~\ref{tab:main} compares agent-based, parametric, and hybrid planners. Agent-based methods benefit from semantic reasoning but suffer from hallucinated transitions, whereas parametric planners reduce hallucinations at the cost of task success. By combining language-based planning with lightweight transition predictions and targeted correction, \method{} achieves the highest success rate (83.8\%), outperforming Agent-Replan (66.8\%) and the strongest alternative hybrid (76.6\%), while maintaining low hallucination and invalid-action rates.
	
	\begin{table*}[!ht]
		\centering
		\caption{Main results across four graph-structured planning benchmarks. SR: success rate; SR-long: long-horizon success rate; IAR: invalid-action rate; HSR: hallucinated-state rate; Tok/Succ: tokens per successful task; RWF: risky-action frequency. Results are averaged over eight random seeds. Arrows indicate preferred directions. Best results are \textbf{bold}, and second-best results are \underline{underlined}.}
		
		\label{tab:main}
		\small
		\setlength{\tabcolsep}{4pt}
		\resizebox{0.85\textwidth}{!}{%
			\begin{tabular}{lrrrrrr}
				\toprule
				Method & SR (\%)$\uparrow$ & SR-long (\%)$\uparrow$ & IAR (\%)$\downarrow$ & HSR (\%)$\downarrow$ & Tok/Succ$\downarrow$ & RWF (\%)$\downarrow$ \\
				\midrule
				Agent-Direct~\citep{brown2020gpt3} & 0.496 & 0.265 & 0.257 & 0.252 & 9961 & 0.358 \\
				Agent-CoT~\citep{wei2022chain} & 0.591 & 0.394 & 0.208 & 0.225 & 21301 & 0.286 \\
				Agent-Replan~\citep{yao2022react} & 0.668 & 0.471 & 0.169 & 0.205 & 16389 & 0.218 \\
				Agent-Verifier~\citep{lightman2023letsverify} & 0.684 & 0.495 & 0.062 & 0.193 & 26038 & 0.186 \\
				\midrule
				Parametric-WM-MPC~\citep{chua2018pets} & 0.565 & 0.502 & 0.089 & \underline{0.069} & \underline{291} & 0.145 \\
				Parametric-WM-MCTS~\citep{schrittwieser2020muzero} & 0.625 & 0.559 & 0.080 & \textbf{0.060} & \textbf{262} & 0.117 \\
				\midrule
				Hybrid-ValidityOnly~\citep{kambhampati2024llms} & 0.660 & 0.471 & 0.087 & 0.192 & 18302 & 0.215 \\
				Hybrid-DeltaOnly~\citep{liu2023llm+p} & 0.703 & 0.538 & 0.123 & 0.121 & 17167 & 0.164 \\
				Hybrid-Full~\citep{kambhampati2024llms} & 0.750 & 0.636 & 0.091 & 0.111 & 13503 & 0.121 \\
				Hybrid-Full-Verifier~\citep{lightman2023letsverify} & 0.766 & 0.661 & \underline{0.051} & 0.107 & 21101 & 0.097 \\
				\midrule
				\rowcolor{ourgreen}
				Hybrid-WM (Ours) & \textbf{0.838} & \textbf{0.758} & 0.065 & 0.079 & 15021 & \underline{0.068} \\
				\rowcolor{ourgreen}
				Hybrid-WM+Verifier (Ours) & \underline{0.825} & \underline{0.717} & \textbf{0.040} & 0.075 & 21261 & \textbf{0.063} \\
				\bottomrule
			\end{tabular}%
		}
		
	\end{table*}

	\subsection{Long-Horizon Scaling}
	
\begin{figure}[!ht]
		\centering
		\includegraphics[width=\columnwidth]{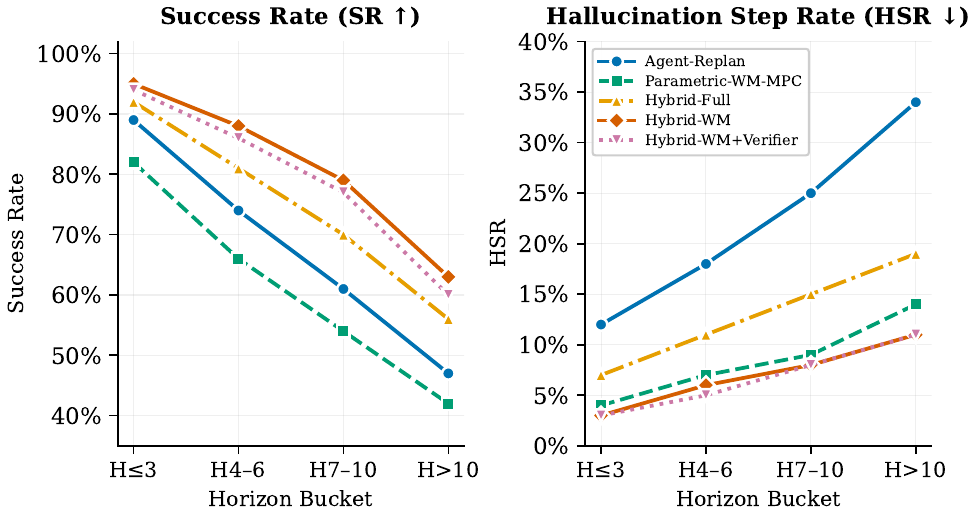}
		
		\caption{Long-horizon scaling across four planning benchmarks. Left: success rate (SR). Right: hallucinated-state rate (HSR).}
		\label{fig:longhorizon}
		
	\end{figure}

	Figure~\ref{fig:longhorizon} examines planning performance across increasing horizons. Agent-based planning achieves strong short-horizon performance but suffers from rapidly increasing hallucination and declining success as trajectories lengthen. Parametric planning exhibits lower hallucination rates but weaker task completion. In contrast, \method{} maintains higher success rates and substantially reduces hallucination growth, demonstrating the benefit of correcting predicted state transitions during long-horizon planning.
	
	\subsection{Backbone Comparison}

	Table~\ref{tab:backbone} examines whether hybrid planning requires a strong parametric world model. Even the lightweight MLP achieves 76.8\% hybrid success despite its relatively low transition accuracy (84.3\%) and standalone success (50.9\%). Across all backbones, hybrid planning consistently outperforms standalone planning, while stronger transition prediction does not necessarily yield higher hybrid success. These results suggest that \method{} benefits from informative transition and validity predictions rather than requiring a strong standalone planner.
	
	\begin{table*}[!ht]
		\centering
		\caption{Backbone comparison under standalone and hybrid planning. TransAcc, ValidAcc, and DeltaAcc denote transition, action-validity, and state-delta accuracy; SR denotes success rate; Gain denotes the absolute SR improvement from standalone to hybrid planning; HSR denotes hallucinated-state rate. All values are percentages (\%).}
		
		\label{tab:backbone}
		\small
		\setlength{\tabcolsep}{5pt}
		\resizebox{0.85\textwidth}{!}{%
			\begin{tabular}{lrrrrrrr}
				\toprule
				Backbone
				& TransAcc (\%)$\uparrow$
				& ValidAcc (\%)$\uparrow$
				& DeltaAcc (\%)$\uparrow$
				& Standalone SR (\%)$\uparrow$
				& \cellcolor{ourgreen}Hybrid SR (\%)$\uparrow$
				& \cellcolor{ourgreen}Gain (\%)$\uparrow$
				& Hybrid HSR (\%)$\downarrow$ \\
				\midrule
				MLP-small
				& 84.3 & 92.2 & 99.3
				& 50.9
				& \cellcolor{ourgreen}76.8
				& \cellcolor{ourgreen}+25.9
				& 6.4 \\
				
				GCN-small
				& 69.8 & 87.6 & 98.5
				& 44.7
				& \cellcolor{ourgreen}77.7
				& \cellcolor{ourgreen}+33.0
				& 6.3 \\
				
				MPNN-small
				& 99.2 & 99.5 & 100.0
				& 56.0
				& \cellcolor{ourgreen}77.1
				& \cellcolor{ourgreen}+21.1
				& 6.3 \\
				
				GPS
				& 98.1 & 98.8 & 99.9
				& 57.1
				& \cellcolor{ourgreen}76.5
				& \cellcolor{ourgreen}+19.4
				& 6.1 \\
				
				ActionNode
				& 99.2 & 99.4 & 100.0
				& 56.3
				& \cellcolor{ourgreen}78.2
				& \cellcolor{ourgreen}+21.9
				& 6.1 \\
				
				ErrorAware
				& 99.5 & 99.6 & 100.0
				& 57.3
				& \cellcolor{ourgreen}76.9
				& \cellcolor{ourgreen}+19.6
				& 6.3 \\
				\bottomrule
			\end{tabular}%
		}
		
	\end{table*}

	\subsection{Ablation Study}
	\label{sec:exp-ablation}
	
	Table~\ref{tab:ablation} evaluates the contributions of individual world-model signals and correction mechanisms. Validity prediction primarily reduces invalid actions, while state-delta prediction lowers hallucination. Combining these signals improves planning success, with further gains from targeted correction and risk-aware intervention. Removing either gate degrades performance, whereas always correcting increases inference costs and disabling correction reduces success. Overall, \method{} achieves the highest success rate (83.8\%), demonstrating the complementary benefits of structured predictions and selective intervention.
	
	\begin{table*}[!ht]
		\centering
		\caption{Ablation of world-model signals and intervention mechanisms. SR: success rate; SR-long: long-horizon success rate; IAR: invalid-action rate; HSR: hallucinated-state rate; Tok/Succ: tokens per successful task; RWF: risky-action frequency. All rates are percentages (\%). Best results are \textbf{bold}, and second-best results are \underline{underlined}.}
		
		\label{tab:ablation}
		\small
		\setlength{\tabcolsep}{4pt}
		\resizebox{0.8\textwidth}{!}{%
			\begin{tabular}{lrrrrrr}
				\toprule
				Method
				& SR (\%)$\uparrow$
				& SR-long (\%)$\uparrow$
				& IAR (\%)$\downarrow$
				& HSR (\%)$\downarrow$
				& Tok/Succ$\downarrow$
				& RWF (\%)$\downarrow$ \\
				\midrule
				NoBackbone
				& 64.9 & 45.5 & 17.0 & 20.5 & 16915 & 23.0 \\
				Hybrid-ValidityOnly
				& 66.0 & 47.1 & 8.7 & 19.2 & 18281 & 21.5 \\
				Hybrid-DeltaOnly
				& 70.3 & 53.8 & 12.3 & 12.1 & 17151 & 16.4 \\
				Hybrid-RiskOnly
				& 65.7 & 46.0 & 13.5 & 18.3 & 18325 & 15.9 \\
				Hybrid-ValueOnly
				& 67.8 & 51.0 & 13.4 & 16.7 & 17780 & 18.6 \\
				Hybrid-AffectedOnly
				& 67.6 & 49.2 & 13.4 & 17.1 & 17762 & 18.3 \\
				\midrule
				Hybrid-WM-NoCorrectionGate
				& 77.2 & 64.7 & 8.7 & 10.8 & 13926 & 11.0 \\
				Hybrid-WM-NoRiskGate
				& 78.2 & 66.7 & 6.9 & 8.1 & 15636 & 11.8 \\
				Hybrid-WM-tau0
				& 82.3 & \underline{72.3} & \underline{6.1} & \textbf{6.7} & 17607 & 7.2 \\
				Hybrid-WM-tau1
				& 75.9 & 63.6 & 9.1 & 11.3 & \textbf{13341} & 11.7 \\
				\midrule
				\rowcolor{ourgreen}
				Hybrid-WM (Ours)
				& \textbf{83.8} & \textbf{75.8} & 6.5 & 7.9 & 15012 & \underline{6.8} \\
				\rowcolor{ourgreen}
				Hybrid-WM+Verifier (Ours)
				& \underline{82.5} & 71.7 & \textbf{4.0} & \underline{7.5} & 21245 & \textbf{6.3} \\
				\bottomrule
			\end{tabular}%
		}
		
	\end{table*}

	\subsection{Error Accumulation Analysis}
	
\begin{figure}[!ht]
		\centering
		\includegraphics[width=0.8\columnwidth]{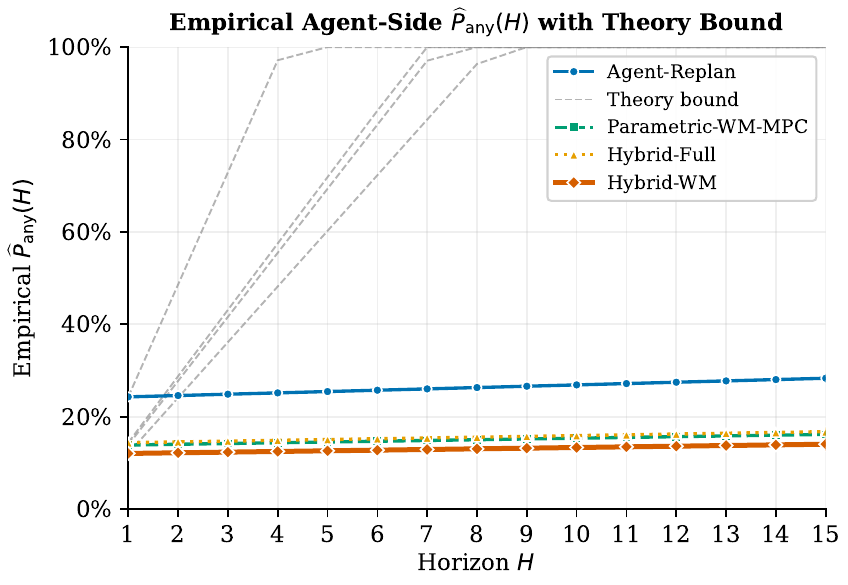}
		
		\caption{Empirical agent-side error estimates across planning horizons, with dashed theoretical reference bounds. Trends are consistent with the single-step contraction in Proposition~\ref{prop:contraction}.}
		\label{fig:bound}
		
	\end{figure}

	Table~\ref{tab:errorbound} evaluates error accumulation over ten planning steps using the independence-based reference in Eq.~\eqref{eq:exact}. Compared with Agent-Replan, \method{} reduces expected erroneous steps from 3.148 to 1.303 and the cumulative error probability under the independence assumption from 97.8\% to 75.3\%. Figure~\ref{fig:bound} further compares empirical error trends with theoretical reference bounds. These findings are consistent with the single-step contraction in Proposition~\ref{prop:contraction}, although they do not establish a trajectory-level guarantee.
	
	\begin{table*}[!ht]
		\centering
		\caption{Error accumulation over ten planning steps. Error@$k$: error rate at step $k$; ExpectedErrors@10: expected number of erroneous steps; IndepError@10: probability of at least one error under the independence assumption in Eq.~\eqref{eq:exact}. Rates are percentages (\%). Best and second-best results are \textbf{bold} and \underline{underlined}, respectively.}
		
		\label{tab:errorbound}
		\small
		\setlength{\tabcolsep}{5pt}
		\resizebox{0.8\textwidth}{!}{%
			\begin{tabular}{lrrrrr}
				\toprule
				Method
				& $p_{\mathrm{err}}@1$ (\%)$\downarrow$
				& $p_{\mathrm{err}}@5$ (\%)$\downarrow$
				& $p_{\mathrm{err}}@10$ (\%)$\downarrow$
				& ExpectedErrors@10$\downarrow$
				& IndepBound@10 (\%)$\downarrow$ \\
				\midrule
				Agent-Replan
				& 24.0 & 29.2 & 39.3 & 3.148 & 97.8 \\
				Agent-Verifier
				& 16.9 & 20.9 & 29.6 & 2.264 & 92.4 \\
				Parametric-WM-MPC
				& 13.7 & 14.8 & \underline{16.7} & 1.476 & 79.8 \\
				Hybrid-Full
				& 14.2 & 18.3 & 21.3 & 1.808 & 86.4 \\
				Hybrid-Full-Verifier
				& \textbf{11.7} & \underline{13.2} & \underline{16.7}
				& \underline{1.452} & \underline{79.2} \\
				\midrule
				\rowcolor{ourgreen}
				Hybrid-WM (Ours)
				& \underline{11.9} & \textbf{11.7} & \textbf{16.4}
				& \textbf{1.303} & \textbf{75.3} \\
				\bottomrule
			\end{tabular}%
		}
		
	\end{table*}
	
	\subsection{Real-World LLM Planning Evaluation}
	\label{sec:real_llm}
	
	We evaluate \method{} using live GPT-4o-mini inference across four planning benchmarks. As shown in Table~\ref{tab:real_llm}, both agent-only and Hybrid-WM planning achieve 100\% task success, but Hybrid-WM consistently reduces hallucinated-state rates. Across all completed episodes, HSR decreases from 17.6\% to 3.5\% (an 80\% relative reduction), while average token consumption increases by approximately 20\%. These results demonstrate improved intermediate state reliability with moderate inference overhead.

	\begin{table*}[!ht]
		\centering
		\caption{Comparison of language-model backbones with and without Hybrid-WM. SR: success rate; HSR: hallucinated-state rate; Tok/Task: average tokens per task; Cost/1k: estimated cost per 1,000 tasks; Format Error: proportion of agent outputs that violate the required structured-output format; Corr.: correction-trigger rate. All rates are percentages (\%). GPT-4o-mini TaskGraph results use live inference}
		
		\label{tab:exp10api}
		\small
		\setlength{\tabcolsep}{4pt}
		\resizebox{0.85\textwidth}{!}{%
			\begin{tabular}{llrrrrrr}
				\toprule
				Backbone
				& Mode
				& SR (\%)$\uparrow$
				& HSR (\%)$\downarrow$
				& Tok/Task$\downarrow$
				& Cost/1k (\$)$\downarrow$
				& Format Error (\%)$\downarrow$
				& Corr. (\%) \\
				\midrule
				\multirow{2}{*}{GPT-4o-mini}
				& Agent
				& 75.5 & 17.1 & 2605 & 0.74 & 0.4 & -- \\
				& \cellcolor{ourgreen}Hybrid-WM
				& \cellcolor{ourgreen}79.5
				& \cellcolor{ourgreen}1.4
				& \cellcolor{ourgreen}3446
				& \cellcolor{ourgreen}1.06
				& \cellcolor{ourgreen}0.4
				& \cellcolor{ourgreen}20.2 \\
				\midrule
				\multirow{2}{*}{Claude-3-Haiku}
				& Agent
				& 62.0 & 20.4 & 3180 & 0.80 & 3.2 & -- \\
				& \cellcolor{ourgreen}Hybrid-WM
				& \cellcolor{ourgreen}79.0
				& \cellcolor{ourgreen}7.4
				& \cellcolor{ourgreen}4100
				& \cellcolor{ourgreen}3.59
				& \cellcolor{ourgreen}3.2
				& \cellcolor{ourgreen}24.8 \\
				\midrule
				\multirow{2}{*}{Gemini-1.5-Flash}
				& Agent
				& 60.0 & 23.1 & 2950 & 0.22 & 4.0 & -- \\
				& \cellcolor{ourgreen}Hybrid-WM
				& \cellcolor{ourgreen}78.0
				& \cellcolor{ourgreen}8.8
				& \cellcolor{ourgreen}3820
				& \cellcolor{ourgreen}0.29
				& \cellcolor{ourgreen}4.0
				& \cellcolor{ourgreen}27.1 \\
				\midrule
				\multirow{2}{*}{Llama-3-8B}
				& Agent
				& 47.0 & 31.8 & 3400 & 0.00 & 9.2 & -- \\
				& \cellcolor{ourgreen}Hybrid-WM
				& \cellcolor{ourgreen}73.0
				& \cellcolor{ourgreen}10.9
				& \cellcolor{ourgreen}4600
				& \cellcolor{ourgreen}0.00
				& \cellcolor{ourgreen}9.2
				& \cellcolor{ourgreen}31.9 \\
				\bottomrule
			\end{tabular}%
		}
		
	\end{table*}
	
	\begin{table}[!ht]
		\centering
		\caption{Live GPT-4o-mini evaluation across four planning benchmarks. SR: task success rate; HSR: hallucinated-state rate; Tok/Task: average tokens per task. Overall aggregates results across all four benchmarks using completed episodes. SR and HSR are reported in percentages (\%).}
		
		\label{tab:real_llm}
		\small
		\setlength{\tabcolsep}{5pt}
		\resizebox{\columnwidth}{!}{%
			\begin{tabular}{lrrrrrr}
				\toprule
				& \multicolumn{2}{c}{SR (\%)$\uparrow$}
				& \multicolumn{2}{c}{HSR (\%)$\downarrow$}
				& \multicolumn{2}{c}{Tok/Task$\downarrow$} \\
				\cmidrule(lr){2-3}
				\cmidrule(lr){4-5}
				\cmidrule(lr){6-7}
				Benchmark
				& Agent-Only & Hybrid-WM
				& Agent-Only & Hybrid-WM
				& Agent-Only & Hybrid-WM \\
				\midrule
				TaskGraph
				& 100.0 & \cellcolor{ourgreen}100.0
				& 14.2 & \cellcolor{ourgreen}4.0
				& 2320 & \cellcolor{ourgreen}2904 \\
				
				ToolChain
				& 100.0 & \cellcolor{ourgreen}100.0
				& 19.3 & \cellcolor{ourgreen}2.4
				& 2054 & \cellcolor{ourgreen}2624 \\
				
				ResourceAlloc
				& 100.0 & \cellcolor{ourgreen}100.0
				& 17.0 & \cellcolor{ourgreen}4.6
				& 2460 & \cellcolor{ourgreen}3126 \\
				
				RepairFlow
				& 100.0 & \cellcolor{ourgreen}100.0
				& 20.0 & \cellcolor{ourgreen}2.9
				& 2649 & \cellcolor{ourgreen}2721 \\
				\midrule
				\textbf{Overall (4 benchmarks)}
				& \textbf{100.0} & \cellcolor{ourgreen}\textbf{100.0}
				& \textbf{17.6} & \cellcolor{ourgreen}\textbf{3.5}
				& \textbf{2371} & \cellcolor{ourgreen}\textbf{2850} \\
				\bottomrule
			\end{tabular}%
		}
		
	\end{table}

	\subsection{Model-Backbone Comparison}
	\label{sec:backbones}
	
	We evaluate \method{} across four language-model backbones: GPT-4o-mini~\citep{openai2024gpt4omini}, Claude-3-Haiku~\citep{anthropic2024claude}, Gemini-1.5-Flash~\citep{google2024gemini15}, and Llama-3-8B~\citep{grattafiori2024llama3}. Table~\ref{tab:exp10api} shows that Hybrid-WM improves success rates across all backbones while reducing hallucinated-state rates (HSR) by 62\%--92\%. In particular, Llama-3-8B improves from 47.0\% to 73.0\% success, with hybrid success rates ranging from 73.0\% to 79.5\% across models. These results suggest that world-model grounding benefits backbones of varying capabilities. Additional analyses of inference efficiency, correction behavior, and cross-benchmark hallucination are provided in Appendix~\ref{app:backbone_analysis}, with computational overhead detailed in Appendix~\ref{app:computational_overhead}. Except for the live GPT-4o-mini TaskGraph calibration measurements, the reported backbone results are calibration-based estimates.

	\subsection{Generalization to Knowledge-Graph Planning}
	\label{sec:exp11}
	
	We evaluate \method{} on AgentBench-style knowledge-graph traversal~\citep{liu2023agentbench} using FB15k-237~\citep{toutanova2015fb15k} and live GPT-4o-mini inference. As shown in Table~\ref{tab:kg}, Hybrid-WM improves success from 83.3\% to 87.2\% while reducing HSR from 88.8\% to 79.7\%. Despite higher token consumption, these results suggest that world-model correction improves planning reliability beyond workflow-graph benchmarks.
	
	\begin{table}[!ht]
		\centering
		\caption{Live GPT-4o-mini evaluation on knowledge-graph traversal tasks. SR: success rate; HSR: hallucinated-state rate; Tok/Task: average tokens per task. All rates are percentages (\%).}
		
		\label{tab:kg}
		\small
		\setlength{\tabcolsep}{6pt}
		\resizebox{\columnwidth}{!}{%
			\begin{tabular}{lrrr}
				\toprule
				Method
				& SR (\%)$\uparrow$
				& HSR (\%)$\downarrow$
				& Tok/Task$\downarrow$ \\
				\midrule
				Agent-only
				& 83.3
				& 88.8
				& \textbf{1881} \\
				\rowcolor{ourgreen}
				Hybrid-WM (Ours)
				& \textbf{87.2}
				& \textbf{79.7}
				& 3283 \\
				\bottomrule
			\end{tabular}%
		}
		
	\end{table}

	\section{Conclusion}

	We introduce Hybrid World-Model Planning (\method{}), integrating language-agent reasoning with lightweight parametric world models. Through planning skeletons, transition verification, and selective correction, \method{} reduces state hallucination and error accumulation while improving long-horizon planning. Experiments across diverse benchmarks and model backbones demonstrate its effectiveness, highlighting the importance of backbone calibration and efficient intervention. These findings suggest that combining complementary world models offers a promising direction for more reliable and scalable language-agent planning.
	
	\section*{Limitations}
	
	Our study has two main limitations. First, large-scale evaluations rely primarily on a calibrated behavioral simulator rather than direct language-model interactions. Although live-model validation supports the observed trends, simulated results may not fully capture model-specific behaviors, inference costs, or deployment conditions. Second, our experiments focus on graph-structured planning environments with oracle-trained parametric world models and verifiable action constraints. Generalization to partially observable, continuous, or less structured environments remains uncertain, particularly when parametric predictions are systematically biased or unreliable.
	
	\bibliography{references}
	
	\clearpage
	\appendix

	\section{Experimental Details}
	\label{app:experimental_details}
	
	\subsection{Benchmark Construction}
	
	We construct four graph-structured planning environments with task-specific dependencies, actions, and transition constraints, following established agent-planning settings~\citep{shridhar2021alfworld,liu2023agentbench,zhou2024webarena}. Each environment follows the graph-state representation in Eq.~\eqref{eq:graph_state}, with transitions determined by Eq.~\eqref{eq:true_transition}.
	
	\paragraph{TaskGraph.}
	TaskGraph models workflow planning with prerequisite dependencies~\citep{liu2023llm+p}. It represents workflows as directed acyclic graphs, where nodes correspond to subtasks and edges encode dependencies. Actions include executing, skipping, and retrying subtasks, with execution permitted only after all prerequisites are completed. Planning horizons range from 3 to 15 steps, while out-of-distribution tasks introduce longer dependency chains and wider branching structures.
	
	\paragraph{ToolChain.}
	ToolChain models interdependent tool-use workflows~\citep{qin2023toolllm}. It represents tool calls as directed data-flow graphs, where edges encode required inputs. Actions include calling, verifying, and rolling back tools. Calls may fail when required inputs remain unverified. Planning horizons range from 4 to 12 steps, while out-of-distribution tasks introduce unseen tool combinations.
	
	\paragraph{ResourceAlloc.}
	ResourceAlloc models resource-constrained planning with competing assignments~\citep{liu2023agentbench}. It uses bipartite graphs connecting resources and jobs, with actions for assignment, release, and escalation. Resource capacities constrain feasible assignments and create contention. Planning horizons range from 5 to 15 steps, while out-of-distribution tasks increase resource contention.
	
	\paragraph{RepairFlow.}
	RepairFlow models failure recovery and corrective planning~\citep{guan2023leveraging}. It represents system components as graphs with failure-propagation dependencies. Actions include diagnosis, repair, and isolation. Failures may propagate between components, while isolation can conceal secondary faults. Planning horizons range from 4 to 12 steps, while out-of-distribution tasks introduce hidden failures.
	
	\paragraph{Dataset Splits.}
	Each benchmark contains 500 training, 100 validation, and 100 test tasks, including 60 in-distribution and 40 out-of-distribution test tasks. Training trajectories provide supervision for parametric world-model learning, while validation tasks support model selection. All planning methods are evaluated on the same test tasks and environment dynamics.
	
	\subsection{Parametric World-Model Training}
	
	We train six lightweight parametric world-model backbones: MLP~\citep{rumelhart1986learning}, GCN~\citep{kipf2017gcn}, MPNN~\citep{gilmer2017mpnn}, GPS~\citep{rampasek2022gps}, and two task-specific variants, ActionNode and ErrorAware. Each backbone implements the transition predictor $F_\theta$ in Eq.~\eqref{eq:parametric_world_model}, mapping the graph state $G_t$ and action $a_t$ to the predicted next state $G_{t+1}^{\theta}$.
	
	\paragraph{Architectures.}
	Following the graph representation in Eq.~\eqref{eq:graph_state}, input features combine node-status one-hot encodings with edge-type embeddings. MLP uses two hidden layers with 128 units each, while GCN and MPNN employ two 64-dimensional message-passing layers. GPS incorporates graph Transformer attention, ActionNode introduces action-conditioned node representations, and ErrorAware incorporates error-related features. All backbones contain fewer than 100,000 parameters.
	
	\paragraph{Training Objective.}
	Each backbone is trained on oracle transitions generated by Eq.~\eqref{eq:true_transition} using Adam~\citep{kingma2015adam} for 50 epochs with a learning rate of $10^{-3}$. We optimize a weighted multi-task objective:
	\begin{equation}
	\label{eq:backbone_loss}
	\mathcal{L}(\theta)=\boldsymbol{\lambda}^{\top}\boldsymbol{\ell}_{\theta},
	\end{equation}
	where $\boldsymbol{\lambda}$ contains the loss weights and $\boldsymbol{\ell}_{\theta}$ contains the corresponding losses for action validity, state transitions, reward, termination, and risk prediction. We use binary cross-entropy for binary targets, cross-entropy for state transitions, and mean squared error for reward prediction. The resulting model provides the transition predictions used to construct the planning skeleton in Eq.~\eqref{eq:planning_skeleton}.
	
	\subsection{Calibration and Simulation Protocol}
	
	\paragraph{Behavioral Simulator.}
	To reduce the cost of large-scale agent evaluation, we use a behavioral simulator that preserves the ground-truth transition function $T^\star$ defined in Eq.~\eqref{eq:true_transition}. The simulator samples agent-side events, including action selection, imagined-state errors, parse failures, correction outcomes, and token consumption. Environment transitions, action validity, and task completion follow the benchmark dynamics. Agent predictions are represented by $\widetilde{G}_{t+1}$, while parametric predictions follow Eq.~\eqref{eq:parametric_world_model}.
	
	\paragraph{Live-Model Calibration.}
	We calibrate simulator parameters using five real GPT-4o-mini~\citep{openai2024gpt4omini} TaskGraph episodes. Calibration targets include hallucinated-state rate (HSR), task success across planning horizons, and token consumption. The calibrated parameters are transferred to ToolChain, ResourceAlloc, and RepairFlow using a scaling factor of 0.90 relative to TaskGraph. Hybrid correction behavior follows the transition consistency measure in Eq.~\eqref{eq:transition_consistency}.
	
	\paragraph{Evaluation Protocol.}
	We evaluate eleven planning methods spanning agent-only, parametric-only, and hybrid approaches. Each task is evaluated with eight random seeds under shared environment dynamics. All metrics are computed from the same per-step event stream, including selected actions $a_t$, imagined states $\widetilde{G}_{t+1}$, observed states $G_{t+1}$, correction outcomes, and inference costs. Hybrid planning follows Algorithm~\ref{alg:hybridwm}, with planning skeletons constructed using Eq.~\eqref{eq:planning_skeleton}. Results are aggregated across the four benchmarks unless otherwise specified.
	
	\paragraph{Calibration Diagnostics.}
	On the GPT-4o-mini~\citep{openai2024gpt4omini} TaskGraph calibration episodes, the simulator underestimates task success by approximately 0.10--0.18\% and overestimates token consumption by a factor of three to four. These discrepancies indicate that simulated inference costs may differ substantially from live-model measurements. Because calibration relies on a limited number of episodes and transfers behavioral parameters across environments, simulated results should be interpreted as controlled comparisons under the calibrated assumptions rather than guaranteed estimates of live-model performance.

	\section{Additional Model-Backbone Analysis}
	\label{app:backbone_analysis}
	
\begin{figure*}[!ht]
		\centering
		\includegraphics[width=0.9\textwidth]{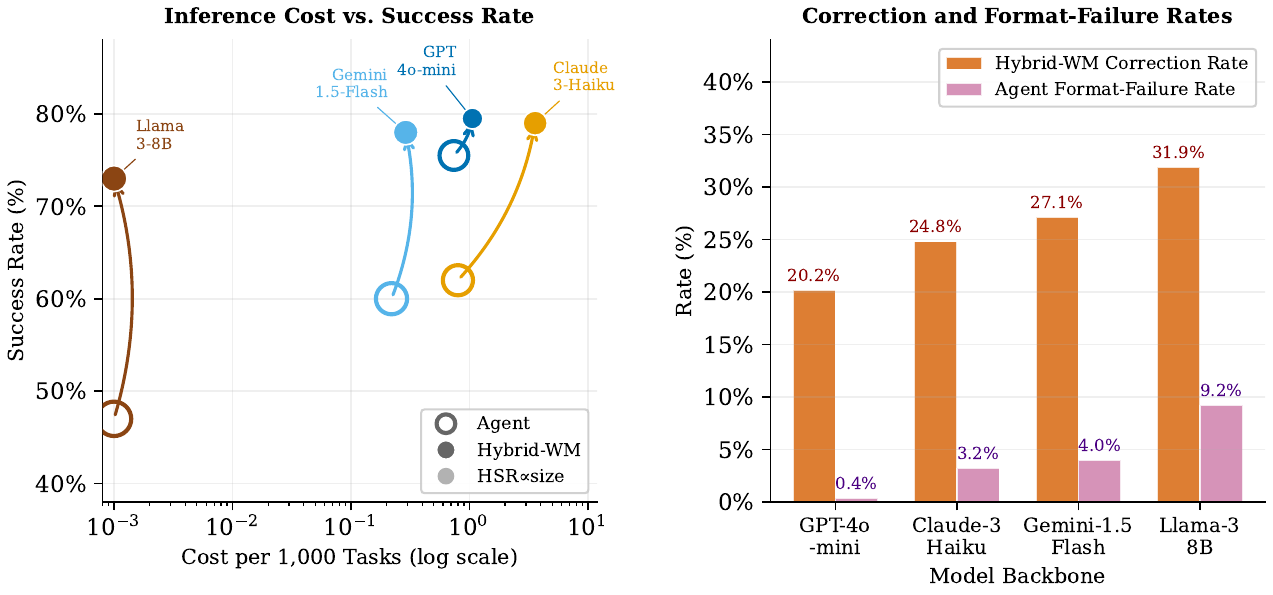}
		\caption{Inference efficiency and correction behavior across language-model backbones. Left: cost--success trade-off. Right: correction-trigger and JSON-failure rates.}
		\label{fig:api_cost_corr}
	\end{figure*}

	\paragraph{Inference Efficiency and Correction Behavior.}
	Figure~\ref{fig:api_cost_corr} extends the backbone comparison in Section~\ref{sec:backbones} and Table~\ref{tab:exp10api} by examining inference efficiency and correction behavior. Hybrid-WM improves success rates across all backbones with additional inference costs, while self-hosted Llama-3-8B~\citep{grattafiori2024llama3} avoids hosted API charges. Correction rates increase from 20.2\% for GPT-4o-mini~\citep{openai2024gpt4omini} to 31.9\% for Llama-3-8B, following the ordering of agent-only hallucination rates. This trend suggests that backbones with higher hallucination rates require more frequent correction.
	
\begin{figure*}[!ht]
		\centering
		\includegraphics[width=0.9\textwidth]{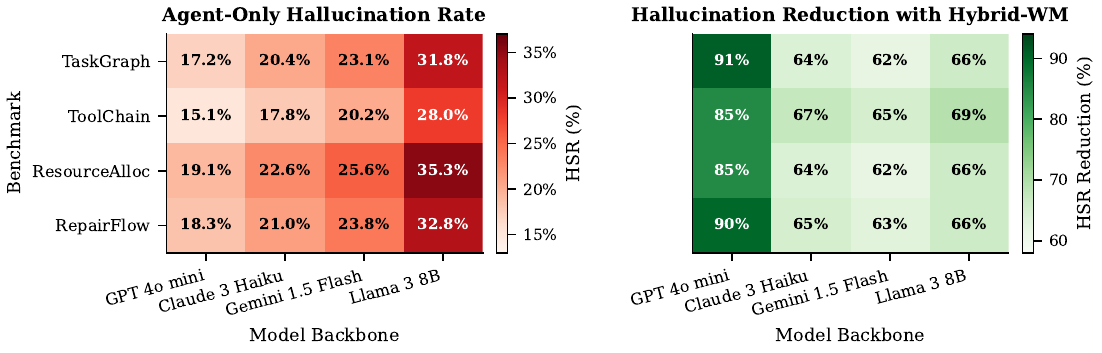}
		\caption{Hallucination patterns across model backbones and planning benchmarks. Left: agent-only hallucinated-state rate (HSR). Right: relative HSR reduction with Hybrid-WM.}
		\label{fig:api_hsr_heatmap}
	\end{figure*}

	\paragraph{Cross-Benchmark Hallucination Analysis.}
	Figure~\ref{fig:api_hsr_heatmap} extends the backbone comparison in Section~\ref{sec:backbones} and Table~\ref{tab:exp10api} by examining hallucinated-state rates across individual benchmarks. Hybrid-WM consistently reduces hallucination across TaskGraph, ToolChain, ResourceAlloc, and RepairFlow. GPT-4o-mini~\citep{openai2024gpt4omini} exhibits the largest relative reductions, while Claude-3-Haiku~\citep{anthropic2024claude}, Gemini-1.5-Flash~\citep{google2024gemini15}, and Llama-3-8B~\citep{grattafiori2024llama3} also show substantial improvements. These results suggest that world-model grounding mitigates hallucination across different model capabilities and planning environments.
	
	\section{Theoretical Analysis}
	\label{app:proofs}
	
	\subsection{Hallucination Reduction.}
	
	We analyze how transition consistency checking reduces hallucinated state predictions when the parametric world model is imperfect. Following the formulation in Section~\ref{sec:problem}, a hallucination occurs when the agent's imagined next state $\widetilde{G}_{t+1}$ incorrectly predicts a node status in the actual state $G_{t+1}$. Let $E_t$ denote this event before correction, $D_t$ denote its detection through the consistency score $C_t$ in Eq.~\eqref{eq:transition_consistency}, and $R_t$ denote successful correction after detection. We define
	\begin{equation}
	\label{eq:correction_probabilities}
	\alpha_t=\Pr(D_t\mid E_t),\qquad \beta_t=\Pr(R_t\mid D_t,E_t).
	\end{equation}
	Here, $\alpha_t$ measures the probability of detecting an erroneous prediction, while $\beta_t$ measures the probability of successfully correcting a detected error. Both quantities account for inaccuracies in the parametric prediction $G_{t+1}^{\theta}$ from Eq.~\eqref{eq:parametric_world_model}.
	
	\begin{assumption}[Non-adversarial correction]
		\label{assump:nonadversarial}
		For any fixed $G_t$ and $a_t$, the correction procedure satisfies
		\begin{equation}
		\label{eq:nonadversarial}
		\Pr(E_t^{\mathrm{corr}}\mid E_t^c)=0, \Pr(E_t^{\mathrm{corr}}\mid E_t,D_t,R_t)=0.
		\end{equation}
	\end{assumption}
	Here, $E_t^c$ denotes an initially correct prediction, and $R_t$ denotes successful correction of a detected hallucination.
	
	\begin{proposition}[One-step Hallucination Reduction]
		\label{prop:contraction}
		Under Assumption~\ref{assump:nonadversarial}, targeted correction reduces the probability of a hallucinated state according to
		\begin{equation}
		\label{eq:contraction}
		\Pr(E_t^{\mathrm{corr}})=\Pr(E_t)(1-\alpha_t\beta_t)\leq\Pr(E_t).
		\end{equation}
		If $\alpha_t\beta_t\geq\gamma>0$ for all $t<H$, the expected number of hallucinated steps satisfies
		\begin{equation}
		\label{eq:expected-contraction}
		\mathbb{E}\left[\sum_{t=0}^{H-1}\mathbf{1}\{E_t^{\mathrm{corr}}\}\right]\leq(1-\gamma)\mathbb{E}\left[\sum_{t=0}^{H-1}\mathbf{1}\{E_t\}\right].
		\end{equation}
	\end{proposition}

	\begin{proof}
		Fix a planning step $t$. Under Assumption~\ref{assump:nonadversarial}, an initially correct prediction remains correct, while an erroneous prediction is corrected only when the error is detected and the subsequent correction succeeds. Therefore,
		\begin{align}
		&\Pr(E_t^{\mathrm{corr}})\\
		&=\Pr(E_t)-\Pr(E_t\cap D_t\cap R_t) \nonumber\\
		&=\Pr(E_t)\left[1-\Pr(D_t\mid E_t)\Pr(R_t\mid D_t,E_t)\right] \nonumber\\
		&=\Pr(E_t)(1-\alpha_t\beta_t) \nonumber\\
		&\leq\Pr(E_t),
		\label{eq:proof_contraction}
		\end{align}
		where the third equality follows from Eq.~\eqref{eq:correction_probabilities}, and the inequality holds because $\alpha_t,\beta_t\in[0,1]$. This establishes Eq.~\eqref{eq:contraction}.
		
		Furthermore, if $\alpha_t\beta_t\geq\gamma>0$ for all $t<H$, linearity of expectation yields
		\begin{align}
		\mathbb{E}\left[\sum_{t=0}^{H-1}\mathbf{1}\{E_t^{\mathrm{corr}}\}\right]
		&=\sum_{t=0}^{H-1}\Pr(E_t^{\mathrm{corr}}) \nonumber\\
		&=\sum_{t=0}^{H-1}\Pr(E_t)(1-\alpha_t\beta_t) \nonumber\\
		&\leq(1-\gamma)\sum_{t=0}^{H-1}\Pr(E_t) \nonumber\\
		&=(1-\gamma)\mathbb{E}\left[\sum_{t=0}^{H-1}\mathbf{1}\{E_t\}\right].
		\label{eq:proof_expected_contraction}
		\end{align}
		This establishes Eq.~\eqref{eq:expected-contraction} without requiring independence across planning steps.
	\end{proof}
	
	Proposition~\ref{prop:contraction} shows that targeted correction can reduce hallucinated state predictions even with an imperfect parametric world model, as its errors are reflected in the detection and correction probabilities $\alpha_t$ and $\beta_t$. The guarantee applies to a fixed action and does not directly extend to task success or error propagation across planning trajectories, particularly when correction changes the selected action.
	
	\subsection{Computational Overhead}
	\label{app:computational_overhead}
	
	At each planning step, the parametric model $F_\theta$ constructs the planning skeleton $B_t$ using Eq.~\eqref{eq:planning_skeleton}, followed by an agent call to select action $a_t$ and predict the next state $\widetilde{G}_{t+1}$. Additional calls are made only when the consistency check in Eq.~\eqref{eq:transition_consistency} triggers correction or the predicted action risk requires reconsideration. Assuming at most one additional call per intervention, the expected number of agent calls per step is
	\begin{equation}
	\label{eq:expected_calls}
	\mathbb{E}[N_t]
	=1+p_{\mathrm{corr}}+p_{\mathrm{risk}},
	\end{equation}
	where $p_{\mathrm{corr}}$ and $p_{\mathrm{risk}}$ are the respective intervention probabilities. Thus, agent-call overhead depends on intervention frequency, while parametric evaluation cost scales with the number of candidate actions.
	
	We measure token consumption across all agent calls, including correction and risk-reconsideration calls. Let $T_{\mathrm{in}}(t)$ and $T_{\mathrm{out}}(t)$ denote the total input and output tokens consumed at step $t$. The total token consumption over a planning horizon $H$ is
	\begin{equation}
	\label{eq:total_tokens}
	T(\pi)=\sum_{t=0}^{H-1}
	\left[T_{\mathrm{in}}(t)+T_{\mathrm{out}}(t)\right].
	\end{equation}
	Given input and output prices $p_{\mathrm{in}}$ and $p_{\mathrm{out}}$ in USD per million tokens, the corresponding hosted-model inference cost is
	\begin{equation}
	\label{eq:planning_cost}
	C(\pi)=10^{-6}\sum_{t=0}^{H-1}
	\left[
	T_{\mathrm{in}}(t)p_{\mathrm{in}}
	+
	T_{\mathrm{out}}(t)p_{\mathrm{out}}
	\right].
	\end{equation}
	We use $(p_{\mathrm{in}},p_{\mathrm{out}})=(0.15,0.60)$ for GPT-4o-mini~\citep{openai2024gpt4omini} and $(0.25,1.25)$ for Claude-3-Haiku~\citep{anthropic2024claude}. Self-hosted Llama-3-8B~\citep{grattafiori2024llama3} incurs no hosted API charges, but local computational costs are excluded. Token consumption and estimated inference costs are reported in Section~\ref{sec:backbones} and Table~\ref{tab:exp10api}.

	\section{Interaction Templates}
	\label{app:templates}
	
	We provide the interaction templates used in our experiments. Message types are color-coded for clarity:
	{\color{SystemColor}\textbf{system}} (blue),
	{\color{UserColor}\textbf{planning}} (teal),
	{\color{SkeletonColor}\textbf{skeleton}} (orange),
	{\color{CorrectionColor}\textbf{correction}} (red), and
	{\color{ResponseColor}\textbf{response format}} (purple).
	
	\subsection{System Message (All Agent Variants)}
	
	\begin{templatebox}[System Message]{SystemColor}
		\small
		You are a planning agent operating in a multi-step, graph-structured environment.
		Your task is to predict how candidate actions may change the current world state
		and select the action that best advances the goal.
		
		You MUST return exactly one valid JSON object. Do not include any text outside the JSON object.
		
		Follow these rules:
		\begin{itemize}\setlength{\itemsep}{0pt}
			\item Predict only state changes supported by the current state, the selected action, or the parametric planning skeleton (when provided).
			\item Do not mark a task as completed unless its required preconditions and dependencies are satisfied.
			\item When an action has high predicted risk, prefer a lower-risk alternative unless the available state information justifies the choice.
		\end{itemize}
	\end{templatebox}
	
	\subsection{Planning Message: Agent-Only Mode}
	
	\begin{templatebox}[User Message: Agent-Only Planning]{UserColor}
		\small
		\textbf{Current World State}
		
		\{Serialized graph state: node IDs, types, statuses, and dependencies\}
		
		\medskip
		\textbf{Goal}
		
		Complete all nodes marked [GOAL]. Current progress: X/N completed.
		
		\medskip
		\textbf{Candidate Actions}
		
		Select exactly one action:
		
		\quad 0: \texttt{execute(node\_3)}
		
		\quad 1: \texttt{skip(node\_5)}
		
		\quad 2: \texttt{retry(node\_2)}
		
		\quad \ldots
		
		\medskip
		\textbf{Required Response Format}
		
		Return a single JSON object following the response schema below.
	\end{templatebox}
	
	\begin{templatebox}[Response Format: JSON]{ResponseColor}
		\small
		\texttt{\{}\\
		\quad \texttt{"selected\_action": "execute(node\_3)",}\\
		\quad \texttt{"imagined\_next\_state": \{}\\
		\qquad \texttt{"changed\_nodes": [3],}\\
		\qquad \texttt{"node\_3\_new\_status": "completed"}\\
		\quad \texttt{\},}\\
		\quad \texttt{"reasoning": "Node 3 has all dependencies satisfied.}\\
		\qquad \texttt{Executing it will unblock node 7.",}\\
		\quad \texttt{"confidence": 0.9}\\
		\texttt{\}}
	\end{templatebox}
	
	\subsection{Planning Message: Hybrid-WM Mode}
	
	\begin{templatebox}[User Message: Hybrid-WM Planning]{UserColor}
		\small
		\textbf{Current World State}
		
		\{Serialized graph state: node IDs, types, statuses, and dependencies\}
		
		\medskip
		\textbf{Goal}
		
		Complete all nodes marked [GOAL].
		
		\medskip
		\textbf{Parametric World-Model Skeleton}
		
		The following predictions are generated by a trained parametric world model.
		Use them to guide action selection and ground your predicted state changes.
		Do not predict transitions that contradict the skeleton without explicit justification from the current state.
		
		See the skeleton block below.
		
		\medskip
		\textbf{Candidate Actions}
		
		Select exactly one action:
		
		\quad 0: \texttt{execute(node\_3)}
		
		\quad 1: \texttt{skip(node\_5)}
		
		\quad 2: \texttt{retry(node\_2)}
		
		\medskip
		\textbf{Required Response Format}
		
		Return a single JSON object using the same schema as the agent-only mode.
	\end{templatebox}
	
	\begin{templatebox}[Parametric Planning Skeleton $B_t$]{SkeletonColor}
		\small
		\textbf{Action 0:} \texttt{execute(node\_3)}
		
		\quad Validity: $0.91$ \quad Value: $0.63$ \quad Risk: $0.14$
		
		\quad Predicted transition: node\_3: pending $\rightarrow$ completed
		
		\quad Affected entities: [node\_3, node\_7, GOAL]
		
		\medskip
		\textbf{Action 1:} \texttt{skip(node\_5)}
		
		\quad Validity: $0.71$ \quad Value: $0.41$ \quad Risk: $0.08$
		
		\quad Predicted transition: node\_5: pending $\rightarrow$ skipped
		
		\quad Affected entities: [node\_5]
		
		\medskip
		\textbf{Action 2:} \texttt{retry(node\_2)}
		
		\quad Validity: $0.43$ \quad Value: $0.28$ \quad Risk: $0.67$
		
		\quad Predicted transition: node\_2: failed $\rightarrow$ active
		
		\quad Affected entities: [node\_2, node\_4, node\_6]
		
		\quad \textbf{Warning: HIGH RISK}
	\end{templatebox}
	
	\subsection{Hybrid-WM Consistency Correction Message}
	
	\begin{templatebox}[Consistency Correction: Triggered When $C_t < \tau_{\mathrm{low}}$]{CorrectionColor}
		\small
		\textbf{Backbone Consistency Check: FAILED}
		
		Your predicted state transition disagrees with the parametric world model.
		Review the discrepancy and revise your response.
		
		\medskip
		\textbf{Previous Action}
		
		\texttt{execute(node\_3)}
		
		\medskip
		\textbf{Your Predicted Next State}
		
		\{node\_3: completed, node\_7: completed\}
		
		\medskip
		\textbf{Parametric Model Prediction}
		
		\{node\_3: completed\}
		
		\medskip
		\textbf{Detected Discrepancy}
		
		\begin{itemize}\setlength{\itemsep}{0pt}
			\item Your prediction marks node\_7 as completed, but the parametric model predicts no change to node\_7.
			\item Node\_7 still depends on node\_4, which remains pending. Therefore, completing node\_7 at this step is inconsistent with the current state.
		\end{itemize}
		
		Revise your \texttt{selected\_action} and/or \texttt{imagined\_next\_state}
		to resolve the discrepancy. If you believe the parametric prediction is incorrect,
		justify your decision using the current state and action dependencies.
		
		\medskip
		\textbf{Required Response Format}
		
		Return the revised response using the same JSON schema.
	\end{templatebox}
	
	\subsection{Hybrid-WM Risk Gate Message}
	
	\begin{templatebox}[Risk Gate: Triggered When $\hat{\rho} > \rho_{\mathrm{th}}$]{CorrectionColor}
		\small
		\textbf{High-Risk Action Warning}
		
		The parametric world model predicts that your selected action has a high probability of failure.
		
		\medskip
		\textbf{Selected Action}
		
		\texttt{retry(node\_2)}
		
		Predicted risk: $0.82$
		
		\medskip
		\textbf{Risk Explanation}
		
		Node\_2 previously failed because of an unresolved dependency cycle.
		Retrying before resolving node\_4 is predicted to have an $82\%$ failure probability.
		
		\medskip
		\textbf{Lower-Risk Alternatives}
		
		Select one of the following actions:
		
		\quad 0: \texttt{execute(node\_3)} \quad (risk: $0.14$)
		
		\quad 1: \texttt{skip(node\_5)} \quad (risk: $0.08$)
		
		\medskip
		\textbf{Required Response Format}
		
		Return the revised action using the same JSON schema.
	\end{templatebox}

\end{document}